%% file: main.tex
\documentclass{article}

\usepackage[preprint]{icml2026}

\usepackage{microtype}
\usepackage{graphicx}
\usepackage{subcaption}
\usepackage{booktabs} 
\usepackage{dblfloatfix} 

\usepackage[hidelinks]{hyperref}

\usepackage{amsmath}
\usepackage{amssymb}
\usepackage{mathtools}
\usepackage{amsthm}
    \usepackage{latexsym}
    \usepackage{amsfonts}
\usepackage{glossaries}
\usepackage{multirow}
\usepackage{array}
    \newcolumntype{P}[1]{>{\centering\arraybackslash}p{#1}}

\usepackage[capitalize,noabbrev]{cleveref}

\icmltitlerunning{TR-PTQ: High-Accuracy Integer-Only Transformer Post Training Quantization via Taylor Region Reformulation}

 \usepackage{breqn}
\usepackage{xcolor}
\usepackage{amsfonts}  
\usepackage{siunitx}
\usepackage{xspace}
\usepackage{glossaries}
\input{newcommands/units}

\input{newcommands/macros}
\input{newcommands/this-glossary}

\begin{document}

\twocolumn[
  \icmltitle{TR-PTQ: High-Accuracy Integer-Only Transformer Post Training Quantization via Taylor Region Reformulation}
\            





\icmlcorrespondingauthor{Eliyahu Levy}{eliyahu.levy@biu.ac.il}
    \begin{icmlauthorlist}
        \icmlauthor{Eliyahu Levy}{biu}
        \icmlauthor{Adam Teman}{biu}
        \icmlauthor{Yoni Pugachov}{biu}
    \end{icmlauthorlist}
    
    \icmlaffiliation{biu}{
        Faculty of Engineering, Bar-Ilan University, Ramat Gan, Israel
    }
    
  \icmlkeywords{Machine Learning, ICML}

  \vskip 0.3in
]



\printAffiliationsAndNotice{}  

\begin{abstract}
Post-training quantization (PTQ) enables efficient deployment, yet transformer architectures remain challenging to quantize due to nonlinear layers. While existing methods attribute accuracy loss to insufficient numerical precision, often necessitating floating-point fallbacks, we demonstrate that degradation is actually driven by specific structural error sources. We find that learned scale parameters in normalization layers and compounded approximations in GELU are the primary error contributors, whereas SoftMax remains inherently robust to aggressive quantization. To address these bottlenecks, we introduce TR-PTQ, a unified integer-only formulation using shared Taylor Region (TR) exponential and logarithm primitives. This approach allows computationally expensive operations, including division and square roots, to be performed entirely in the log-domain via standard integer arithmetic. Combined with a calibration-free, outlier-aware optimization for LayerNorm parameters, our method eliminates the need for floating-point hardware units for nonlinearities, achieving less than 1.5\% absolute accuracy degradation across vision and language benchmarks.
\end{abstract}


\section{Introduction}

\Gls{PTQ} enables efficient model deployment by bypassing the costs of retraining and extensive calibration. Despite these benefits, transformer architectures present significant challenges to quantization due to their heavy reliance on nonlinear layers. Prevailing approaches typically attribute the resulting accuracy loss to insufficient numerical precision. Consequently, existing methods often resort to complex calibration procedures, specialized observers, or mixed-precision execution with floating-point fallbacks.

In this work, we challenge the assumption that nonlinear layers fundamentally require high-precision arithmetic. We demonstrate that accuracy degradation under aggressive per-tensor \PTQ is driven not by a general lack of precision, but by specific, identifiable structural error sources. By isolating these sources, we show that many nonlinear components can tolerate aggressive quantization when handled appropriately, while others fail due to specific parameter interactions rather than inherent numerical sensitivity.

We first identify that in normalization layers, such as \layernorm, quantization error is dominated by the learned scale parameters ($\gamma$) rather than by input activations. The heavy-tailed distributions of these parameters interact poorly with standard min-max and histogram-based observers, often inducing larger accuracy degradation than errors introduced by mean or root-mean-square (RMS) estimation. For nonlinear activation functions, such as \GELU, we identify an approximation-on-approximation effect where surrogate implementations compounded with quantization lead to amplified errors.

In contrast, we find that \softmax is inherently robust under aggressive quantization. By rounding intermediate activations prior to exponentiation, \softmax can be accurately implemented using a compact lookup table with as few as eight entries, significantly smaller than the commonly used 256-entry designs. This robustness results from logarithmic scaling and normalization, which preserve relative relationships while suppressing absolute magnitude errors. This renders inaccuracies in low-probability elements largely inconsequential.

Building on these insights, we propose a unified, integer-only optimization framework for \PTQ of transformer nonlinear layers. By restructuring nonlinear computations and their associated parameters into low-precision formulations, our method eliminates the need for calibration data for nonlinearities, specialized observers, or floating-point arithmetic units. 

\vspace{3px}\noindent Our specific contributions include the following:
\begin{itemize}
    \item We demonstrate that accuracy degradation in post-training quantization of transformer nonlinearities is driven by a small set of structural error sources rather than by insufficient numerical precision. This finding challenges the prevailing assumption that nonlinear layers fundamentally require high-precision floating-point arithmetic.

    \item We introduce a unified, integer-only formulation for transformer nonlinear operations built around shared Taylor Region exponential and logarithm primitives. This enables hardware-efficient implementations of \softmax, \GELU, and \layernorm without floating-point fallback.
    
    \item We identify and address two under-emphasized bottlenecks in \PTQ: (i) compounded approximation effects in nonlinear activations such as \GELU, and (ii) the dominance of learned scale parameters ($\gamma$) in normalization layers. We propose a simple, calibration-free outlier-aware optimization to effectively mitigate these errors.
       
    \item We validate the proposed approach on vision and language transformers, demonstrating consistent performance under aggressive quantization while significantly simplifying the \PTQ pipeline.

\end{itemize}

\section{Background and Related Work} \label{sec:related_work}

\subsection{Quantization of Nonlinear Layers in Transformers} \label{sec:quantization paradigms}

While a large body of prior work focuses on quantizing linear layers in transformers through pruning and post-training quantization~\cite{frantar2022gptq,pmlr-v202-xiao23c,fastertransformer}, these approaches typically leave nonlinear components such as \softmax, \GELU, and normalization layers in floating point. Consequently, they are largely orthogonal to this work.

Early efforts to quantize nonlinear layers primarily targeted \softmax. Softermax~\cite{softermax} proposed a low-level implementation based on replacing the exponential base from $e$ to $2$, enabling more efficient low-precision computation. I-BERT~\cite{kim2021ibert} later introduced fully integer-only transformers by employing second-order polynomial approximations for nonlinear functions. This strategy was extended to vision transformers by I-ViT~\cite{li2023vit}, which mitigates inter-channel activation variance through power-of-two base conversions.

FQ-ViT~\cite{lin2021fqvit} adopts a \PTQ approach to avoid the high training cost incurred by I-BERT and I-ViT. It reuses the I-BERT \softmax implementation while introducing a power-of-two factor (PTF) to mitigate accuracy degradation in \layernorm. Subsequent work extended integer-only quantization to additional nonlinear components. Zhang et al.~\cite{zhang2023practical} proposed a range-constrained quantization scheme for \GELU, while Sun et al.~\cite{11270824} introduced a segmented \GELU approximation and decomposed the affine parameters $\gamma$ and $\beta$ to enable a fully quantized \layernorm computation.

Although the aforementioned works succeed in approximating nonlinear layers, they typically rely on computationally expensive components, such as dividers, square-root units, or high-precision arithmetic. In contrast, other works such as~\cite{SOLE2023_ICCAD, ARNorm2025_CSCC, Park2025_CSCC, 11270824} propose solutions free of dividers and multipliers, trading off accuracy for lower implementation costs. These involve methods like \emph{E2Softmax} and the Approximate Log-based Division (ALDivision) scheme, which replaces both the exponential and the division in \softmax with only shift, addition, and multiplexer operations. This approach approximates $e^{X}_{Q}$ in a manner similar to~\cite{li2023vit} and performs the division entirely in the $\log$-domain. QUARK~\cite{QUARK2025_ICCAD} reformulates nonlinear computations entirely in the $\log$ domain by applying second-order polynomials. It further introduces reorder-based group quantization to mitigate the inter-channel activation heterogeneity arising from nonlinear layers such as \layernorm and \GELU.

A parallel line of work emphasizes hardware-efficient nonlinear computation. Approaches such as SOLE~\cite{SOLE2023_ICCAD} and related designs~\cite{ARNorm2025_CSCC,Park2025_CSCC} replace exponentiation and division with shift-and-add operations using approximate log-domain formulations. This enables multiplier- and divider-free implementations. QUARK~\cite{QUARK2025_ICCAD} also leverages log-domain reformulation but adds complexity through reorder-based group quantization.

Despite these advances, many existing methods either rely on per-channel or group-wise quantization, which incurs increased hardware and control-path complexity, or they retain nonlinear layers in floating point. This leaves open the question of whether per-tensor quantization can be effectively applied to nonlinear operations without sacrificing accuracy.

\section{Integer-only Taylor Region Approximation for Nonlinear Functions}
\label{sec:tr_format}

In this section, we present unified, integer-only approximations for the exponential ($\exp$) and natural logarithm ($\ln$) functions using a Taylor-based integer representation. These nonlinear functions serve as central components in essential deep-learning operations, such as \softmax and \GELU. Consequently, these approximations form the fundamental building blocks for the complete operators described in the subsequent section.

The proposed Taylor Region $\exp$ (\texp) and $\ln$ (\tln) approximations operate \textbf{entirely in the integer domain} using shift and multiply operations. This design is highly efficient for quantized hardware accelerators and low-power processing units. Both functions are expressed in terms of a base evaluation point, or \textit{anchor} $a_i$ (derived from the leading bits of the input), and a small offset calculated based on the residual distance ($\Delta$) of the input from the evaluation point. The evaluation leverages truncated Taylor series and power-of-two shifts instead of complex division logic. This shared structure simplifies the reuse of arithmetic logic and lookup tables across both nonlinearities. Following the introduction of the methods for calculating $\exp$ and $\ln$, we will demonstrate the full integer-only computations for \softmax, \GELU, and \layernorm.

\subsection{Taylor Region-Based Approximation of the Exponential Function (TR-exp)}
\label{sec:tr_exp}

The proposed approximation of the exponential function (\texp) relies on bounding the input domain, quantizing the values, and utilizing a small \lut with pre-calculated evaluation points to pin a low-order Taylor series approximation.

The first step requires bounding the inputs to ensure they are non-positive. This is achieved by subtracting the maximum value $x_\max$ from the original inputs $x$, yielding $\hat{x}=x-x_\max \leq 0$. This effectively bounds the output within the range $\exp(\hat{x})\in (0, 1]$.

To address quantization, we interpret the binary layout of the quantized input values as three distinct fields: a 1-bit \textit{sign}, an $l$-bit \textit{evaluation point} (\textit{evp}), and a $k$-bit \textit{fraction} (\textit{frac}). For a $b$-bit quantization scheme where $b=l+k$, the structure is defined as follows\footnote{Since all inputs are non-positive, the sign value is implicit, meaning only $b$ bits are strictly required for storage.}:
\begin{equation}
    \label{eq:binary-mapping}
  \bigl[\underbrace{\text{\textit{sign}}}_{1\text{-bit}}
  \;\vert\;
  \underbrace{\underbrace{\text{\textit{evp}}}_{l\text{-bits}}
  \;\vert\;
  \underbrace{\text{\textit{frac}}}_{k\text{-bits}}\bigr]}_{b\text{-bits}}.
\end{equation}
We then quantize the inputs to $2^b$ discrete levels by scaling by $2^k$ while clamping the lower bound. Since the exponent of inputs with very large negative magnitudes approaches zero, we define:
\begin{equation}\label{quntized number}
    \hat{x}_q = 
    \begin{cases} 
      \Bigl\lfloor\hat{x} \cdot 2^{k} \Bigr\rfloor  & \text{if } \hat{x} > -2^{b}, \\ 
      -2^{b}                              & \text{if }  \hat{x} \leq -2^b
    \end{cases}.
\end{equation}

\begin{figure}[t]
    \centering
    \includegraphics[width=\columnwidth]{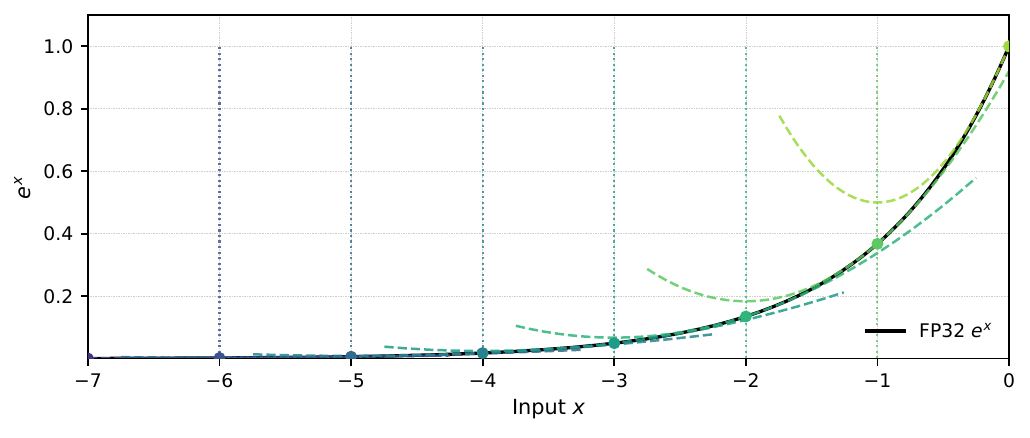}
    \vspace{-0.6em}
    \caption{
    Taylor Regions for $e^x$.
    Vertical dotted lines denote anchor points; dashed curves show local Taylor
    approximations centered at each anchor.
    }
    \vspace{-0.8em}
    \label{fig:taylor-regions}
\end{figure}

Next, we use the \textit{evp} field to extract evaluation points for the exponent calculation. This is performed by pre-calculating $2^l$ exponents $\exp{(a_i)}$, where $a_i \in \{0,\dots,2^{l}\}$ is the anchor. These values are stored in a small \lut and represent the nearest integer to which the Taylor series is pinned. The proposed format allocates $l+1$ bits: a single sign bit and $l$ index bits that address the \lut.
The motivation for this approach is illustrated in \figref{fig:taylor-regions}, which plots the $\exp$ function with the Taylor approximation around the anchor points.

All tabulated values correspond to strictly negative exponents; therefore, the \lut contains exactly $2^{l}$ entries. The value associated with zero input, $\exp(0)=1$, is handled as a designated non-\lut case encoded directly by the sign bit. This eliminates the need to reserve a \lut location for the zero exponent while preserving a uniform indexing structure for all negative inputs.

Recall that the $n$-th order Taylor approximation of the exponential function around an anchor point $a$ is given by:
\begin{equation}
    \exp(x) \approx e^{a}\sum_{i=0}^{n}\frac{(x-a)^{i}}{i!},
    \label{eq:taylor-exp_n}
\end{equation}
where $x$ is the input and $a$ denotes the chosen evaluation point. For $n=2$, the exponential of a quantized and range-limited input $\hat{x}_q$ can be approximated as:
\begin{equation}
    \exp(\hat{x}_q) \approx \text{LUT}[\textit{evp}]\left( 1 + \Delta + \frac{\Delta^{2}}{2} \right),
    \label{eq:tr-exp-wrong}
\end{equation}
where $\text{LUT}[\textit{evp}]$ stores the precomputed anchor value corresponding to the index \textit{evp}, and $\Delta = \hat{x}_q - (\textit{evp} \ll k)$ represents the local offset from the anchor. However, all factors in \eqref{eq:tr-exp-wrong} must be scaled to $b$-bit precision using the scaling factor $2^k$. The corrected integer-only expression is:
\begin{equation}
    \exp{(\hat{x}_q)} \approx \text{LUT}[\textit{evp}]
    \left( 1\ll k + \Delta + (\Delta^2\gg (k+1)) \right)\gg k.
    \label{eq:tr-exp-right}
\end{equation}
Here, the quadratic term is rescaled by right-shifting by $k$, and the division-by-two is achieved with an additional bit-shift (i.e., $\gg (k+1)$). The result is rescaled by right-shifting the final product by $k$.

An additional benefit of this approach is that $\Delta$ only requires $k$ bits, since the $l$ MSBs of $\hat{x}_q$ are equal in each segment \textit{evp} and therefore cancel out during subtraction. Consequently, the calculation of $\Delta$ demands low bit precision and can be calculated once, then efficiently pipelined for the calculation of each additional order of the Taylor series. This structure also allows precision to be tuned based on accuracy requirements, as shown in \secref{sec:experiments}.


\subsection{Shift-based Taylor Region Natural Logarithm Approximation (TR-ln)}
The previous subsection introduced our Taylor Region-based approximation for the exponential function (\texp), a core operator in deep-learning workloads such as \softmax. Building on this, we observe that by employing a lightweight implementation of the natural logarithm (the complementary function of the exponential), we can perform additional nonlinear operations directly in the $\log$-domain with negligible accuracy degradation. This advantage becomes particularly pronounced when the $\exp$ operator requires higher precision. In this subsection, we present a low-cost logarithm approximation suitable for integration into computations such as \softmax and \layernorm.

Exploiting the sub-linear behavior of the natural logarithm over a bounded interval, we replace the full $\ln(\cdot)$ computation with a first-order (linear) approximation. To further reduce approximation error, the input domain is divided into piecewise-linear segments $[a, b]$. The segment boundaries can be selected efficiently in hardware by using the position of the leading 1-bit of $x$ together with the position of the next most significant bit. These bits determine the interval $2^{\lfloor \log_2 x \rfloor}$ in which the value resides, effectively selecting the appropriate linear region. As the segments widen, the function slope decreases, thereby ensuring numerical stability.

Under this construction, the logarithm of the quantized input $x_q$ is approximated as:
\begin{equation}
    \begin{aligned}
      \ln{x_q} = m(x_q-x_{q0}) + \ln{2^a}, 
      \label{eq:linear_ln_apprx}
    \end{aligned}
\end{equation}
where $a$ denotes the position of the leading-one bit, $x_{q0}$ is the corresponding power-of-two anchor point determined by that leading bit, and the slope $m$ is selected according to the chosen segment. The slope can be expressed as:
\begin{equation}
    \begin{aligned}
      m = \frac{\ln{2^{a+1}} - \ln{2^{a}}}{2^{a+1} - 2^{a}}
      & = \frac{\ln{2}}{2^a}
      \label{eq:slop_linear_ln_apprx}
    \end{aligned}
\end{equation}

Under these definitions and substitutions, \eqref{eq:linear_ln_apprx} becomes:
\begin{equation}
    \begin{aligned}
      \ln{x_q} & = 2^{-a}\cdot \ln2 (x-2^{a}) + a\ln2\\ 
       & = \ln2(2^{-a}x-1) + a\ln2\\
        & = \ln2(2^{-a}x+a-1).
      \label{eq:slop_linear_ln_apprx2}
    \end{aligned}
\end{equation}
Note that $\ln2$ can be approximated numerically as $\ln 2 \approx 0.5 + 0.125 + 0.0625$. Since these coefficients are powers of two, the associated multiplications can be realized using shift-and-add operations rather than full multipliers. Thus, defining $k = 2^{-a} x + (a - 1)$, we obtain:
\begin{equation}
    \begin{aligned}
         k \cdot \ln 2 \approx (0.5 + 0.125 + 0.0625)\, k,
    \end{aligned}
\end{equation}
and therefore:
\begin{equation}
    \begin{aligned}
        \ln{x_q} \approx (k \gg 1) + (k \gg 3) + (k \gg 4).
        \label{eq:linear_ln_apprx_in_shift_format}
    \end{aligned}
\end{equation}

\section{Hardware-Efficient TR Calculation of Nonlinear Deep Learning Operators}
\label{sec:nonlinear_operators}

In this section, we introduce a unified, hardware-efficient methodology for approximating nonlinear functions based on the Taylor Region approach developed in Section \ref{sec:tr_format}. We first demonstrate the approach on \softmax and then extend it to \GELU and \layernorm, enabling a fully quantized inference pipeline for vision and language tasks. These proposed computations reduce the cost of operations such as division, square root, and exponentiation into simple additions, shifts, and integer multiplications. This dramatically improves efficiency on resource-constrained hardware by removing the need for floating-point arithmetic units, dividers, and other complex hardware blocks.

\subsection{Taylor Region \softmax Approximation (\tsoft)}

Given an input vector \(x=[x_1,x_2,\dots,x_{n}]\) of length \(n\), the \softmax operator maps each element to a probability \(p_i\in[0,1]\) such that \(\sum_{i=1}^{n} p_i = 1\). Formally:
\begin{equation}
  \operatorname{SoftMax}(x_i)=
  \frac{\exp(x_i)}{\sum_{j=1}^{n}\exp(x_j)} .
  \label{eq:softmax}
\end{equation}

To ensure that the exponentials do not overflow in a quantized representation, the first step is to subtract the maximum input value from all inputs, $\hat{x}=x-x_\max$, as described in \secref{sec:tr_exp}. This simple modification guarantees numerical stability while leaving the output distribution unchanged. Following bounding and scaling, \eqref{eq:softmax} can be rewritten as:
\begin{equation}
    \operatorname{SoftMax}(\hat{x}_{q,i})=\frac{\exp(\hat{x}_{q,i})}{\sum_{j=1}^{n}\exp(\hat{x}_{q,j})},
\end{equation}
where $\hat{x}_{q,i}$ are the quantized and bounded input values.

To compute the denominator $S \triangleq \sum_{j=1}^{n} \exp(\hat{x}_{q,j})$, we exploit the fact that at least one input satisfies $x_j = x_{\max}$, implying $\hat{x}_{q,j} = 0$ and $\exp(\hat{x}_{q,j}) = 1$. Consequently, $S \geq 1$ and $\ln S \geq 0$. This allows us to evaluate the reciprocal $S^{-1}$ using the \tln and \texp operators:
\begin{equation}
    S^{-1} = \exp(-\ln S),
\end{equation}
where the exponent $-\ln S$ is non-negative, as required by the \texp approximation.

\begin{figure*}
    \centering
    \includegraphics[
        width=\textwidth,
        trim=0.2cm 0.5cm 0.2cm 0.1cm,
        clip
    ]{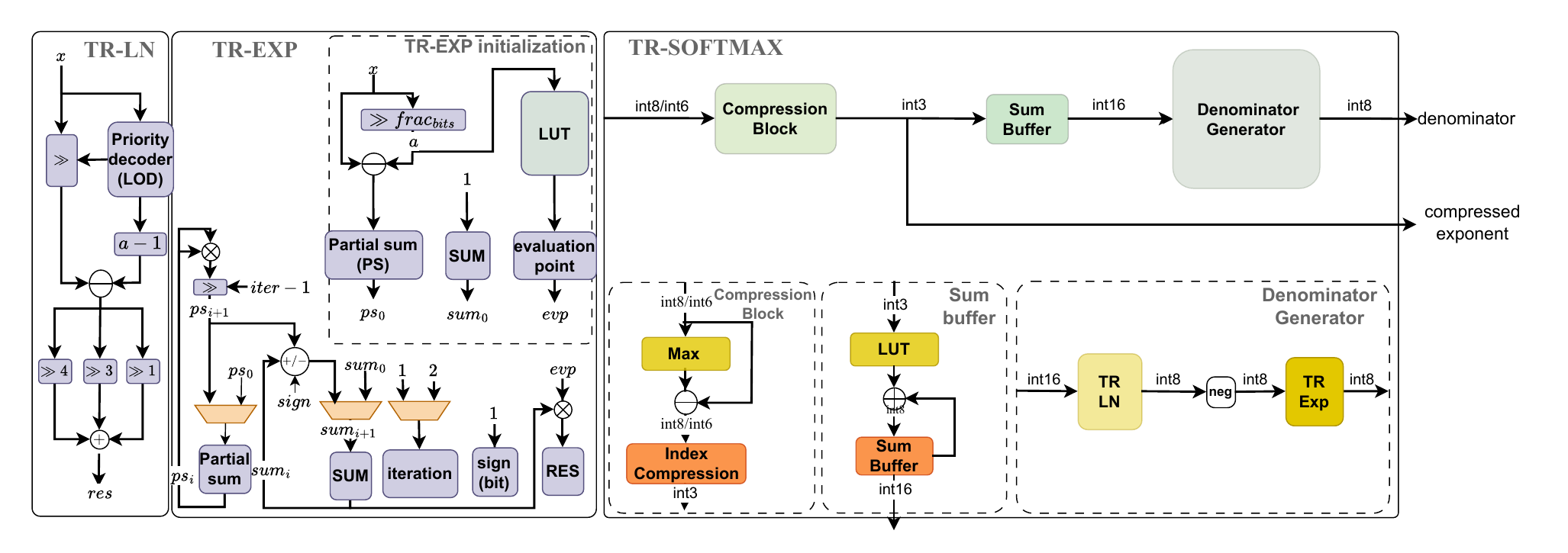}
    \caption{
    Block-level overview of \tln, \texp, and \tsoft.
    The same $\exp$ and $\log$ primitives are reused across attention,
    normalization, and activation functions to enable a unified and
    hardware-efficient implementation.
    }
    \label{fig:tr-nonlinear-blocks}
\end{figure*}

To obtain the final \softmax value, we multiply $S^{-1}$ by $\exp(\hat{x}_{q,i})$ and apply a right shift by $k$ bits for rescaling. The complete \tsoft computation is therefore:
\begin{equation}
    \begin{aligned}
        \operatorname{TR\text{-}SoftMax}[\hat{x}_{q,i}]
        =
        s_f \cdot 
        \operatorname{TR\text{-}\exp}[\hat{x}_{q,i}] \gg k,
    \end{aligned}
    \label{eq:tr-softmax}
\end{equation}
where the normalization factor $s_f$ is computed once per vector:
\begin{equation}
    s_f=\operatorname{TR\text{-}\exp}\!\left(
        -\,\operatorname{TR\text{-}\ln}\!\left[
            \sum_{j=1}^{n} 
            \operatorname{TR\text{-}\exp}(\hat{x}_{q,j})
        \right]
    \right).
\end{equation}

All operations involving additions, shifts, multiplications, and \lut lookups are implemented at $b$-bit precision. This enables an efficient, fully integer arithmetic realization. \figref{fig:tr-nonlinear-blocks} illustrates the implementation of \texp, \tln, and \tsoft.

\subsection{Taylor Region GELU Approximation (TR-GELU)}
Another primary nonlinear function used in various deep learning models is the \acrfull{GELU}~\cite{gelu2016}. \GELU combines linear and nonlinear behaviors smoothly, often yielding better convergence than ReLU in both NLP and vision models. It is defined as:
\begin{equation}
  \operatorname{GELU}(x)
  = x\,\Phi(x)
  = x\,\frac{1}{2}\Bigl(1+\operatorname{erf}\left(\tfrac{x}{\sqrt{2}}\right)\Bigr),
  \label{eq:gelu_exact}
\end{equation}
where $\Phi(x)$ is the Gaussian \cdf and $\operatorname{erf}(\cdot)$ is the error function. 

Direct evaluation of $\operatorname{erf}(\cdot)$ is prohibitively expensive in dedicated hardware. A common workaround replaces $\Phi(x)$ with a scaled sigmoid:
\begin{equation}
  \operatorname{GELU}(x)\approx x\,\sigma(\alpha x),
  \quad
  \sigma(z)=\frac{1}{1+e^{-z}},
  \label{eq:gelu_sigmoid}
\end{equation}
where $\alpha\approx1.702$ is chosen to best fit the exact \GELU curve in a least-squares sense~\cite{gelu2016}. This reduces complexity to a single multiply and a lookup-friendly logistic approximation.

An alternative exploits the ``\softmax-style'' identity~\cite{li2023vit}:
\begin{equation}
  \sigma(x)
  = \frac{1}{1+e^{-x}}
  = \frac{e^{0}}{e^{0}+e^{-x}}
  = \frac{e^{-x_{\max}}}{e^{-x_{\max}} + e^{\,x - x_{\max}}},
  \label{eq:gelu_softmax_form}
\end{equation}
where $x_{\max}=\max(0,x)$ ensures that the exponentials are applied to non-positive numbers for numerical stability. 

The \GELU approximation in \eqref{eq:gelu_sigmoid} and \eqref{eq:gelu_softmax_form} allows the use of the same \texp and \tln operators employed earlier for \tsoft. However, this sigmoid-based form is only reliable when the model was trained with the same approximation. For production models trained with the exact \GELU in \eqref{eq:gelu_exact}, substituting \eqref{eq:gelu_sigmoid} at inference introduces a noticeable accuracy degradation.

To address this limitation, we replace the global scaling factor \(\alpha\) with a region-dependent parameterization. The scaling factors \(\{\alpha_i\}\) are computed offline by explicitly minimizing the mean-square error between the approximate and exact \GELU functions over their corresponding intervals. In our implementation, we use ten uniformly spaced segments covering the range \([-5,5]\).

Due to the odd symmetry of the \GELU function, \(\operatorname{GELU}(-x) = -\operatorname{GELU}(x)\), the optimal scaling factors for negative inputs are obtained by mirroring the positive regions. As a result, only half of the scaling coefficients need to be stored in practice. This reduces the \lut to four entries without loss of accuracy, further minimizing memory overhead.

\subsection{Taylor Region Layer Normalization Approximation (\tnorm)}
The final nonlinear operation required for a fully-quantized \vit is \layernorm. Introduced by Ba et al.~\cite{ba2016layer}, \layernorm rescales each feature vector according to:
 \begin{equation}   
    \begin{aligned}
    &\operatorname{LayerNorm}(x_i)=\gamma\,\frac{x_j-\mu}{\sqrt{\sigma^{2}+\epsilon}}+\beta, \\
    &\mu=\tfrac1n\sum_{j=1}^n x_j,\;
    \sigma^{2}=\tfrac1n\sum_{i=1}^n(x_j-\mu)^2,\;
     \end{aligned}
\end{equation}
where $n$ is the vector size, and $\gamma$ and $\beta$ are trainable parameters. Because \(\mu\) and \(\sigma^{2}\) span a wide range, low-bit quantization is fragile. Consequently, most accelerators maintain a high-precision path via standalone dividers~\cite{kim2021ibert,lin2021fqvit,li2023vit} or 32-bit ``fake-FP32'' blocks~\cite{fastertransformer}.
Iterative square-root (\(\sqrt{\cdot}\)) schemes, such as Newton-Raphson~\cite{crandall2001computational}, reduce the latency to only a few iterations, but still invoke a divider every cycle.

\textbf{Log-Domain Reformulation.} Similar to the approach for \tsoft, we reduce the hardware requirements of the division and square-root by employing \texp and \tln:
\begin{equation}
\begin{aligned}
    \frac{1}{\sqrt{\sigma^{2}+\epsilon}}
    &= \exp\!\left[ \ln (\sigma^{2}+\epsilon)^{-\frac{1}{2}} \right] \\
    &\approx \operatorname{TR\text{-}exp}\!\left[-\tfrac12\,\operatorname{TR\text{-}ln}\!\bigl(\sigma^{2}+\epsilon\bigr)\right].
\end{aligned}
\label{eq:tr-sqrt}
\end{equation}

\begin{figure}[t]
    \centering
    \includegraphics[
        width=\columnwidth,
        trim=0.1cm 0.05cm 0.01cm 0.063cm,
        clip
    ]{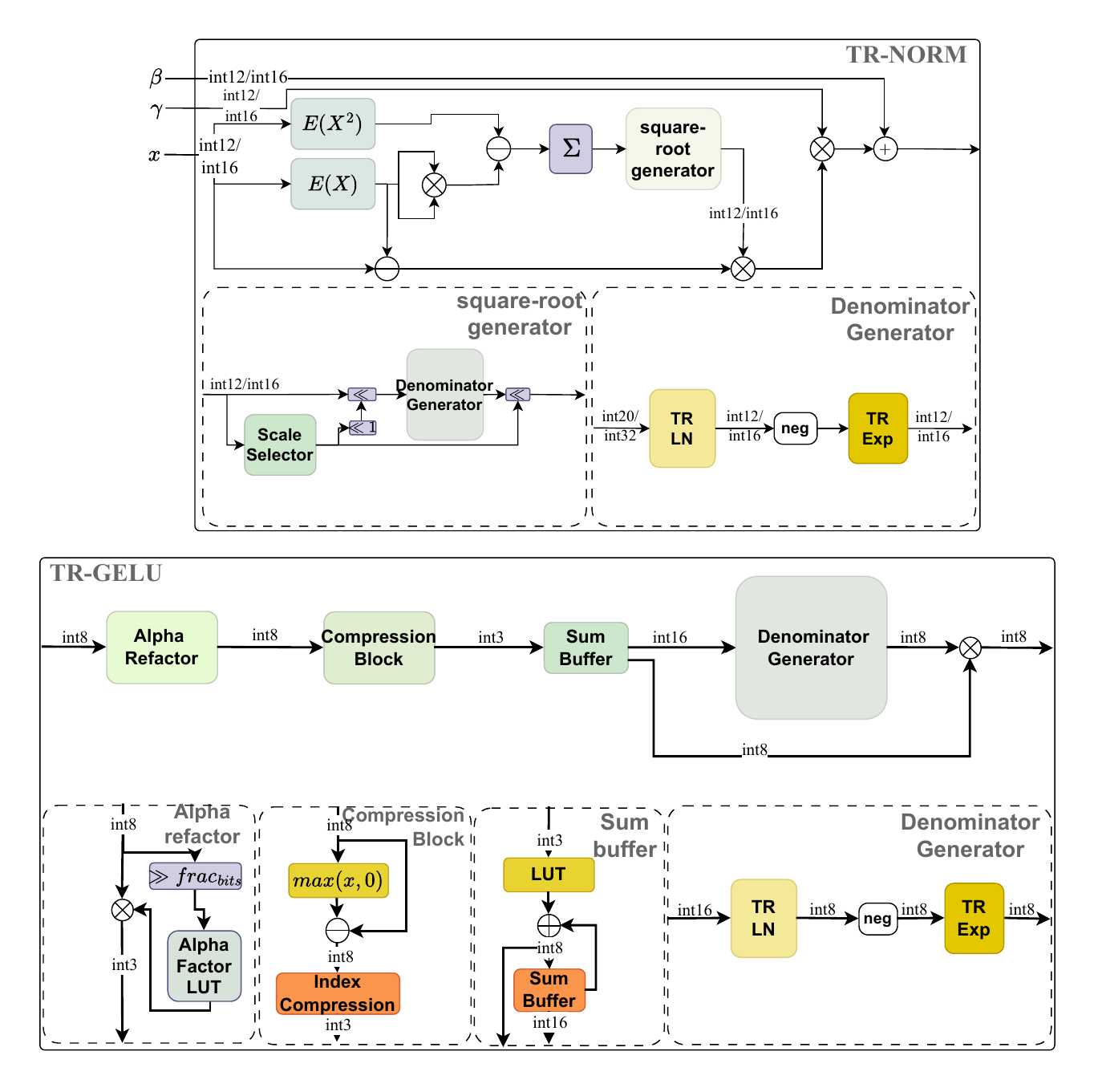}
    \caption{
    Block-level overview of the proposed \tnorm (top) and \tgelu (bottom)
    operators, illustrating their integration within the low-precision
    nonlinear processing pipeline.
    }
    \label{fig:tr-norm-gelu}
\end{figure}

A difficulty emerges when $\sigma^{2} + \epsilon < 1$, as the input to the logarithm becomes negative, resulting in reduced numerical fidelity in low-bit quantized formats. Addressing this region requires a computation mechanism whose precision can be selectively increased. To accommodate this, the $\exp$ \lut is expanded to include positive entries. Although this modification doubles the table size, it preserves the architectural structure and introduces only negligible hardware overhead. \figref{fig:tr-norm-gelu} provides the schemes for \tgelu and \tnorm.

\textbf{Outlier-Aware Scale Optimization.} As shown in \figref{tab:optimized_gamma_effect_glue}, a significant source of accuracy degradation in low-precision \layernorm arises from quantization of the scale parameter $\gamma$, whose distribution is highly skewed and outlier-dominated.
In some configurations, this effect leads to severe accuracy loss, while in others it manifests as consistent two-digit percentage degradation.
Direct min--max quantization therefore induces excessively wide dynamic ranges, collapsing most values into a small number of quantization levels.
While prior work such as Quark~\cite{QUARK2025_ICCAD} mitigates this issue via group-wise quantization, we adopt a complementary approach that optimizes a global scaling factor for $\gamma$ using a lightweight line search.
Specifically, we employ a variance-normalized RMSE objective inspired by EasyQuant~\cite{wu2020easyquantposttrainingquantizationscale} to balance sensitivity to extreme outliers with accurate representation of typical values.
The RMSE term penalizes large deviations caused by high-magnitude $\gamma$ entries, while normalization by the tensor variance prevents degenerate solutions in which most elements collapse to a narrow dynamic range, yielding a stable and well-conditioned scale under aggressive quantization.

\section{Experiments and Evaluation}
\label{sec:experiments}

\begin{table}[b]
    \centering
    \caption{Comparison of Top-1 accuracy (\%) across different \vit models and quantization methods.}
    \label{tab:accuracy_evaluation}
    \renewcommand{\arraystretch}{1.1}
    \resizebox{1.05\columnwidth}{!}{%
        \begin{small}
        \begin{sc}
        \begin{tabular}{lcccccc}
        \toprule
        \textbf{Method} & \textbf{DeiT-T} & \textbf{DeiT-S} & \textbf{DeiT-B} & \textbf{Swin-T} & \textbf{Swin-S} & \textbf{Swin-B} \\
        \midrule
        Full FP32 (baseline)       & 72.21 & 79.85 & 81.85 & 81.35 & 83.20 & 83.60 \\
        \midrule
        FQ-ViT\tiny{~\cite{lin2021fqvit}} & 71.61 & 79.17 & 81.20 & 80.51 & 82.71 & 82.97 \\
        Zhang\tiny{~\cite{zhang2023practical}} & 71.08 & 78.49 & 80.74 & 80.03 & 82.29 & 82.67 \\
        SOLE (INT8)                & 71.07 & 78.89 & 81.12 & 80.14 & 82.60 & 82.79 \\
        QUARK\tiny{~\cite{QUARK2025_ICCAD}} & 71.29 & 79.40 & 81.52 & 81.06 & 82.81 & 85.03\textsuperscript{\textdagger} \\
        \textbf{\tvit (Ours)}      & \textbf{71.39} & \textbf{79.27} & \textbf{81.40} & \textbf{80.70} & \textbf{82.80} & \textbf{83.18} \\
        \bottomrule
        \end{tabular}
        \end{sc}
        \end{small}
        
    } 
    \tiny{\textsuperscript{\textdagger}QUARK reports 85.27\% accuracy on Swin-B when evaluated against its own FP32 baseline.}
\end{table}

\subsection{Implementation Details}
\label{sec:implememntation}

We implemented a fully-quantized, hardware-efficient \vit model, \tvit, utilizing the \texp, \tln, \tsoft, \tgelu, and \tnorm primitives proposed in Sections~\ref{sec:tr_format} and \ref{sec:nonlinear_operators}. \tvit is evaluated on widely adopted \vit architectures, including DeiT~\cite{touvron2021training} and Swin Transformer~\cite{liu2021swin}, using the ImageNet-1k dataset~\cite{deng2009imagenet}. To ensure a comprehensive analysis across varying model capacities, we test three configurations: tiny (T), small (S), and base (B). For post-training calibration, we sample 100 images from the training set and report accuracy on the full 50,000-image validation set.

All experiments were conducted on NVIDIA T4 GPUs. Our implementation utilizes Hugging Face libraries with streaming dataset support. 
While this may introduce minor latency, it significantly lowers the barrier for conducting large-scale experiments.

\subsection{Quantization Configuration}
\label{sec:quant config}
Per-tensor uniform quantization was applied across all major components. Linear projections and matrix multiplications are quantized to 8-bit precision for both inputs and weights. 

The \softmax (\tsoft) and \GELU (\tgelu) computations, which require the evaluation of exponential and logarithmic functions, are implemented using 8-bit integer arithmetic. The specific precision requirements associated with the exponent across various layers are analyzed in the ablation studies (Section \ref{sec:disscussion}).

Layer normalization (\tnorm) is realized using 12-bit arithmetic. This increased precision is necessary to maintain numerical stability during the computation of the mean, variance, and affine parameters ($\gamma$ and $\beta$). Although \tnorm operates at a higher bit-width than other functional units, it eliminates the need for floating-point hardware while ensuring robust behavior across diverse model architectures.

\subsection{Error_Analysis}

\begin{figure}[t]
    \centering
    \includegraphics[width=\columnwidth]{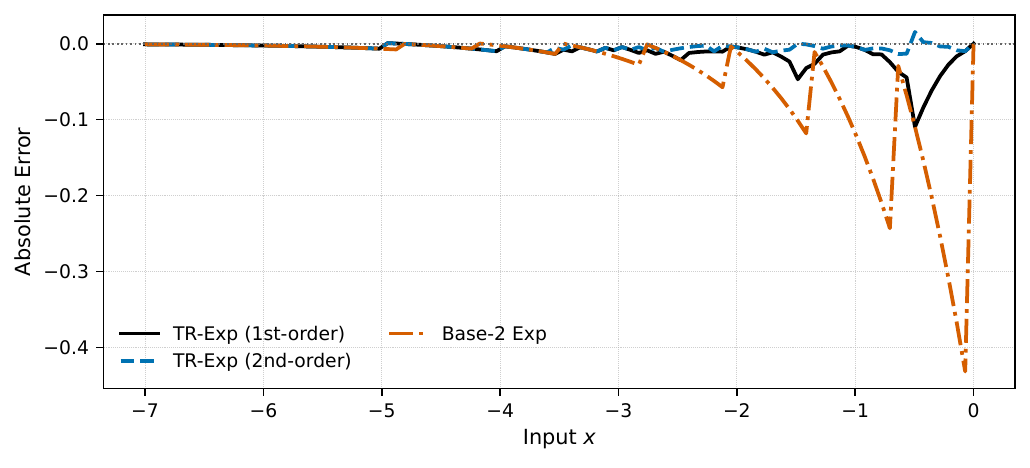}
    \vspace{-0.6em}
    \includegraphics[width=\columnwidth]{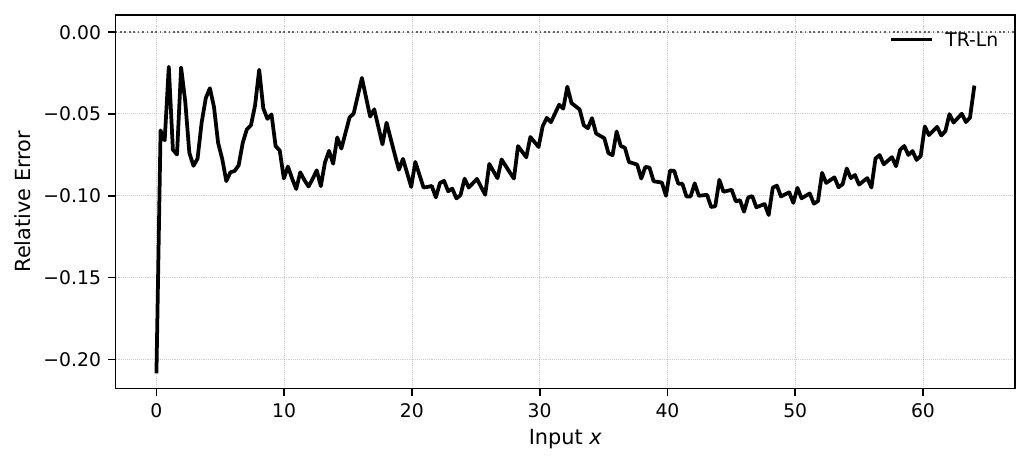}
    \caption{
    Approximation error of the proposed TR nonlinear operators (top: \texp, bottom: \tln)
    under 8-bit integer arithmetic.
    }
    \label{fig:tr-error-core}
\end{figure}

The first-order TR-Exp approximation exhibits bounded absolute error below $0.1$ over the evaluated range, as shown in \figref{fig:tr-error-core}. Since the exponential is subsequently normalized (e.g., $1/\sum \exp(\cdot)$), this error is further attenuated in practice, resulting in accurate $\log$-domain division and inverse square-root operators (see Appendix~\ref{apndx:tr-norm-tr-div}). The TR-Ln approximation shows larger error (up to $0.2$) only for very small inputs ($x \approx 10^{-2}$), while remaining accurate for larger values. Notably, the proposed formulation replaces conventional double-width division with simple shift and multiplication operations, enabling both division and inverse square root using only two same-width multiplications.

\subsection{Accuracy Evaluation}
\label{sec:acc eval}
\tvit is benchmarked against the full-precision (FP32) baseline and recent state-of-the-art \PTQ methods, including FQ-ViT~\cite{lin2021fqvit}, SOLE~\cite{SOLE2023_ICCAD}, QUARK~\cite{QUARK2025_ICCAD}, and the method proposed by Zhang et al.~\cite{zhang2023practical}, which integrates I-BERT's integer \GELU with a constrained quantization scheme.

\tblref{tab:accuracy_evaluation} summarizes the Top-1 accuracy on ImageNet. Our method is among the top performer across all configurations. Compared to the FP32 baseline, the accuracy degradation remains less than 1\% for the majority of models, a notable achievement given the aggressive integer-only quantization applied to all nonlinear layers.

\section{Ablation Study}
\label{sec:disscussion}

To understand the effect of various design decisions on \tvit, we conduct ablation studies across three axes: (1) \softmax approximation, (2) \GELU quantization, and (3) \layernorm precision.

\subsection{\softmax Robustness}
\label{sec:disscussion-softmax}

Replacing the full-precision \softmax operation with the proposed 8-bit \tsoft approximation results in less than 1\% accuracy degradation on vision tasks and modest relative degradation across GLUE benchmarks. The maximum relative drop remains below 4\%, with most tasks exhibiting less than 3\% degradation that causes by the linear layers.

Despite the widespread perception that exponential functions are highly sensitive to approximation, we find that the \softmax operation is surprisingly robust to quantization noise. This robustness stems from two key factors. First, the normalization step rescales each exponential relative to the sum, thereby attenuating the impact of individual approximation errors. Second, the attention mechanism primarily relies on preserving the ordering and relative magnitude of dominant logits rather than exact probability values; contributions from strongly negative logits (e.g., $e^{-8} \approx 3\times10^{-4}$) are negligible and can be coarsely approximated without affecting the output.

To further stress the design, we evaluate increasingly aggressive exponential approximations. Using a zero-order approximation in Q4.4 format (comprising 1-bit sign, 3-bit integer, and 4-bit fractional components) has minimal impact on accuracy. This suggests that the attention mechanism is largely invariant to the exact precision of the exponential tail, provided the relative order of magnitudes is preserved.In contrast, the max-subtraction stage ($\hat{x} = x - x_{max}$) is significantly more sensitive to quantization noise. As shown in \tblref{tab:accuracy_vs_bit_precision}, reducing the precision of this stage to 4–6 bits leads to larger degradation, particularly in compact architectures like DeiT-T. This sensitivity arises because max-subtraction acts as a global shift; errors in $x_{max}$ propagate to every element in the SoftMax input, potentially shifting the distribution into regions where the Taylor approximation is less accurate or where the exponential values saturate near zero. These results indicate that performance degradation is primarily driven by max-subtraction, rather than by inaccuracies in the exponential approximation itself.

\begin{table}[t]
    \centering
    \caption{Model Accuracy Across Different Bit-Precisions}
    \label{tab:accuracy_vs_bit_precision}
    \renewcommand{\arraystretch}{1.1}
    \setlength{\tabcolsep}{4pt}
    \resizebox{\columnwidth}{!}{
    \begin{small}
    \begin{sc}
    \begin{tabular}{lcccccc}
    \toprule
    \textbf{Method} & \textbf{DeiT-T} & \textbf{DeiT-S} & \textbf{DeiT-B} & \textbf{Swin-T} & \textbf{Swin-S} & \textbf{Swin-B} \\
    \midrule
    FP32\tiny{ (baseline)}    & 72.21 & 79.85 & 81.85 & 81.35 & 83.20 & 83.60 \\
    \midrule
    4-bit\tiny{ (rounding)}        & 68.11 & 77.21 & 80.53 & 79.23 & 80.81 & 81.28 \\
    6-bit\tiny{ (rounding)}        & 70.64 & 78.61 & 81.29 & 80.28 & 82.67 & 82.97 \\
    8-bit\tiny{ (rounding)}        & 70.84 & 78.50 & 81.21 & 80.34 & 82.73 & 82.98 \\
    \textbf{8-bit\tiny{ (1st order)}}   & \textbf{71.39} & \textbf{79.27} & \textbf{81.40} & \textbf{80.70} & \textbf{82.80} & \textbf{83.18} \\
    \bottomrule
    \end{tabular}
    \end{sc}
    \end{small}
    }
\end{table}

\subsection{\GELU Optimization}
\label{sec:disscussion-gelu}

Our design choice to adopt the proposed compression was inspired by the structural similarity between the attention mechanism and the feed-forward (MLP) layers, both of which function as key-value memory systems. Based on our finding that attention does not require high-precision representations, we hypothesized that a similar compression strategy would apply to MLP layers.

We push this idea further by evaluating \GELU under an aggressive mixed-precision configuration: activation inputs and outputs use 8 bits, while the exponential used in the sigmoid approximation is computed using only 4-bit precision (denoted as 8/4/8). This setting represents the lowest-precision exponential approximation considered in our study.

Results indicate that smaller models are particularly sensitive. For example, on DeiT with this precision and a global scaling factor $\alpha = 1.702$, the accuracy degradation is approximately 1.15\%. In this regime, \GELU becomes the dominant source of error. However, the proposed region-based scaling recovers nearly all of this residual loss, demonstrating that fine-grained control of the activation function is necessary to fully exploit aggressive quantization, as shown in Table~\ref{tab:gelu_alpha_segmentation}.

\begin{table}
    \centering
    \caption{Impact of \GELU implementation on model accuracy. Only the \GELU nonlinearity is quantized; all other operations remain in floating point.}
    \label{tab:gelu_alpha_segmentation}
    \renewcommand{\arraystretch}{1.1}
    \setlength{\tabcolsep}{6pt}
    \resizebox{0.85\columnwidth}{!}{%
        \begin{small}
        \begin{sc}
        \begin{tabular}{lc}
        \toprule
        \textbf{Configuration} & \textbf{Accuracy (\%)} \\
        \midrule
        FP32 baseline (all layers) & 72.21 \\
        Integer \GELU ($\alpha=1.702$) & 71.07 \\
        \textbf{Integer \GELU (segmented $\alpha_i$)} & \textbf{72.13} \\
        \bottomrule
        \end{tabular}
        \end{sc}
        \end{small}
    }
\end{table}

\subsection{Layer Normalization Analysis} 
\label{sec:disscussion-lnorm}

Across the operational range, the proposed log-domain Taylor approximation remains stable and accurate, with precision controlled by a small number of \texp iterations (at most two in all evaluated configurations), yielding a favorable accuracy-cost trade-off.
Although worst-case degradation under full quantization is observed in \tblref{tab:optimized_gamma_effect_glue}, ablation analysis indicates that it is primarily driven by the linear layers.
In contrast, per-tensor quantization of nonlinear operators introduces only marginal degradation, while the dominant source of error in \layernorm arises from the scale parameter $\gamma$, motivating our $\gamma$-specific treatment.

\begin{table}[t]
    \centering
    \caption{Performance comparison (\%) across GLUE tasks under different quantization configurations.}
    \label{tab:optimized_gamma_effect_glue}
    \renewcommand{\arraystretch}{1.1}
    \setlength{\tabcolsep}{3pt}
    \resizebox{\columnwidth}{!}{
    \begin{small}
    \begin{sc}
    \begin{tabular}{lccccccc}
        \toprule
        \textbf{Config} & \textbf{CoLA} & \textbf{QNLI} & \textbf{RTE} & \textbf{MNLI} & \textbf{QQP} & \textbf{SST-2} & \textbf{MRPC} \\
        \midrule
        FP32 Base & 53.38 & 91.54 & 72.56 & 84.57 & 90.91 & 92.89 & 89.81 \\
        \midrule
        All Quant & 51.27 & 88.30 & 70.76 & 81.94 & 90.07 & 92.78 & 88.43 \\
        Piecewise & 54.80 & 88.30 & 70.76 & 82.75 & 90.79 & 92.88 & 89.84 \\
        Excl Linear & 55.91 & 90.99 & 71.48 & 83.48 & 90.79 & 93.23 & 90.97 \\
        \midrule
        Norm (No Opt) & 2.95 & 65.17 & 55.23 & 53.48 & 70.55 & 86.01 & 38.75 \\
        \textbf{Norm (Ours)} & \textbf{53.18} & \textbf{91.05} & \textbf{71.48} & \textbf{83.65} & \textbf{90.82} & \textbf{92.63} & \textbf{90.01} \\
        \bottomrule
    \end{tabular}
    \end{sc}
    \end{small}
    }
\end{table}

\section{Conclusions}

We demonstrate that key nonlinear components in transformers can be implemented
efficiently at low bit precision using per-tensor quantization while preserving
high accuracy.
Our analysis reveals that accuracy degradation is driven by a small number of
sensitive operations, motivating targeted optimizations that recover most of
the lost performance without increasing hardware complexity.
These results emphasize the importance of selectively allocating precision to
enable efficient low-precision inference.


\bibliography{ref/references.bib}  
\bibliographystyle{icml2026}

\appendix
\onecolumn 
\icmltitle{Appendix for: TR-PTQ: High-Accuracy Integer-Only Transformer Post Training Quantization via Taylor Region Reformulation} 

\counterwithin{figure}{section}
\counterwithin{table}{section}

\appendix
\section{Algorithmic and Hardware Details}

\subsection{Hardware-Efficient Taylor Implementation}
The primary motivation of the Taylor-Region (TR) formulation is to reduce
transcendental functions to operations that map efficiently onto fixed-point
hardware. By aligning the $Q_{I.K}$ representation with the Taylor expansion
centers, the proposed design avoids high-latency arithmetic units and enables
compact, deterministic implementations.

\paragraph{Rounding-based Taylor centering.}
To minimize approximation error, we center the Taylor expansion at the
nearest integer value $a = \mathrm{round}(x)$ rather than at
$a = \lfloor x \rfloor$. In a $Q_{I.K}$ fixed-point representation, this
rounding is implemented by inspecting the most significant fractional bit
$x[k-1]$. Specifically,
\[
a =
\begin{cases}
\lfloor x \rfloor + 1, & \text{if } x[k-1] = 1, \\
\lfloor x \rfloor,     & \text{otherwise}.
\end{cases}
\]
This choice ensures that the residual $\Delta x = x - a$ lies within
$[-0.5, 0.5]$, reducing the maximum approximation error by a factor of two
relative to floor-based centering.

\paragraph{Addition-free first-order evaluation.}
The first-order Taylor term $(1 + x - a)$ is computed without a full-width
adder by exploiting the structure of the fractional field
$f = x[k-1:0]$. Depending on the rounding outcome, the expression simplifies
to:
\[
1 + x - a =
\begin{cases}
\{1, f\}, & \text{if } a = \lfloor x \rfloor, \\
\{0, f\}, & \text{if } a = \lfloor x \rfloor + 1.
\end{cases}
\]
In hardware, this replaces a $b$-bit addition with a single multiplexer
controlled by $x[=k-1]$, significantly reducing area and power consumption.

\paragraph{Logic-based quadratic approximation.}
Under rounding-based centering, the residual satisfies $|\Delta x| \le 0.5$,
and the quadratic term $\Delta x^2 / 2$ is bounded within $[0, 0.125]$. For a
$Q_{4.4}$ configuration, a 4-bit signed residual maps to a 2-bit output.
Exploiting this narrow range, the second-order term is implemented using
combinational logic synthesized via Boolean minimization, rather than a
general-purpose multiplier.

By realizing the quadratic correction as a fixed 4-to-2 bit Boolean function,
the evaluation can be performed in parallel with the first-order term. As a
result, the exponential $\exp(x)$ is reduced to:
(i) a small lookup table (e.g., 16 entries) for the anchor $e^{a}$,
(ii) a multiplexer for the first-order linear term, and
(iii) combinational logic for the second-order curvature.
This structure enables single-cycle evaluation of nonlinear operators such as
\softmax and GELU, while maintaining a low hardware footprint.

\paragraph{Fixed-point representation and Taylor-region mapping.}
We use signed two's-complement fixed-point numbers in the $Q_{I.K}$ convention, where $I$ counts integer bits \emph{including the sign bit} and $K$ counts fractional bits (total width $I{+}K$). For \softmax, we apply standard numerical stabilization via max-subtraction, $x' = x - x_{\max}$ with $x_{\max}=\max(x)$, which enforces $x' \le 0$. Hence, although $Q_{4.4}$ nominally spans $[-8,\,7.9375]$, only the non-positive half is exercised in practice, i.e., $x' \in [-8,\,0]$. Equivalently, the integer field contains one sign bit and three magnitude integer bits. In the Taylor-Region (TR) formulation, the integer component selects the region (evaluation point), while the fractional component is the local residual used by higher-order terms. Our primary configurations are $Q_{4.4}$ (8-bit), $Q_{4.2}$ (6-bit), and $Q_{4.0}$ (4-bit).

\subsection{Preprocessing and numerical stabilization.}
Given an input vector $x$, we compute the per-configuration maximum
$x_{\max}=\max(x)$ and form the stabilized vector $x' = x - x_{\max}$ so that
all entries are non-positive. Importantly, the max-reduction and subtraction
are performed \emph{in the target bit-precision} (8-bit, 6-bit, or 4-bit,
respectively), using the corresponding $Q_{I.K}$ format. The stabilized values
$x'$ are then used as inputs to the exponential evaluation.

\paragraph{Zero-order exponential via a compact LUT.}
To stress-test robustness, we evaluate a zero-order approximation in which the fractional bits are discarded (i.e., we keep only the integer component of $x'$). The exponential is then computed using a $2^{I}$-entry lookup table indexed by the integer value (e.g., 16 entries for $I=4$):
\[
\mathrm{LUT}[a] = \mathrm{round}\!\left(e^{a}\cdot\left(2^{b-1}-1\right)\right),
\]
where $b$ is the output bit-width. Values with $a<-6$ are clamped to zero to avoid underflow. This provides a lightweight alternative to implementing a full 8-bit exponential table (256 entries), while keeping the evaluation hardware-efficient.

\paragraph{Accuracy vs.\ bit precision.}
Table~\ref{tab:accuracy_vs_bit_precision} reports Top-1 accuracy across DeiT and Swin models. Relative to FP32, 4-bit rounding exhibits the largest degradation, with a drop of $1.32$--$4.10$ points (median $2.35$). In contrast, 6-bit and 8-bit rounding reduce the loss to $0.53$--$1.57$ (median $0.85$) and $0.47$--$1.37$ (median $0.83$) points, respectively. 1st order configuration further improves accuracy, limiting the drop to $0.40$--$0.82$ points (median $0.52$).

\subsection{TR-Norm and TR-Div}
\label{apndx:tr-norm-tr-div}
\paragraph{Numerical behavior of TR-Div and TR-Norm.}
The evaluation of TR-Div remains numerically stable under the considered
Transformer workloads. Since the denominator is constrained to be
$\ge 1$ - either through the addition of a stabilization constant $\epsilon$
or by properties of the attention score normalization-the corresponding
logarithm is non-negative. As a result, division in the log domain yields
values bounded by unity, which in turn constrains the subsequent exponential
evaluation to a limited dynamic range. This bounded regime is well suited to
low-order Taylor approximations; empirically, a second-order polynomial
provides the best accuracy-complexity trade-off.

\begin{figure}[t]
    \centering
    \includegraphics[
        width=0.7\columnwidth,
        trim=0 0 0 0,
        clip
    ]{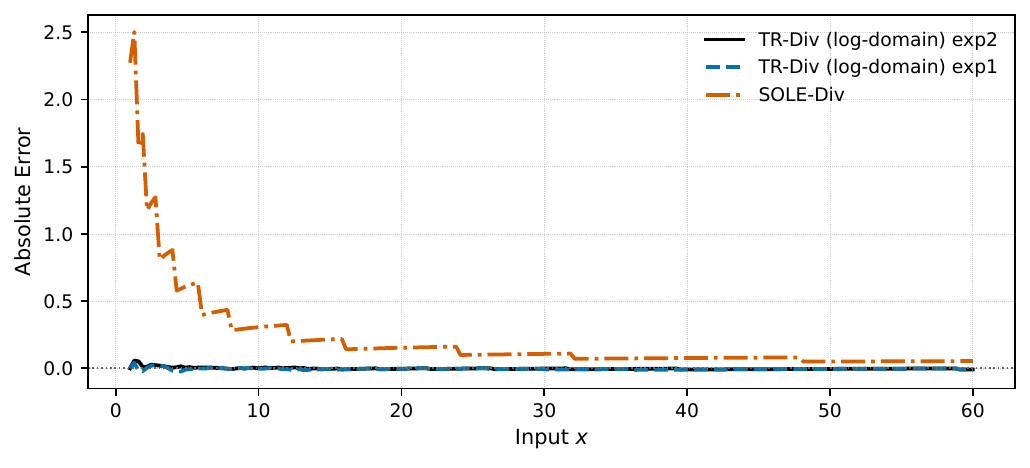}
    \caption{
    Comparison of division noise between the proposed method and a
    low-cost comparable solution (SOLE).
    The plot reports the absolute error with respect to a high-precision
    reference across the evaluated input range.
    }
    \label{fig:division_noise_sole}
\end{figure}

Figure~\ref{fig:division_noise_sole} compares TR-Div against a shift-based
division method~\cite{SOLE2023_ICCAD}. The results indicate that TR-Div
exhibits lower numerical error across the evaluated ranges, which we
attribute to the avoidance of coarse quantization effects introduced by
bit-shift operations.

In contrast, TR-Norm is more sensitive to quantization due to its broader
dynamic range. Unlike TR-Div, the logarithmic inputs encountered in
normalization layers may take both positive and negative values, requiring a
wider representation to preserve accuracy. We therefore adopt a 12-bit
configuration as a balance between numerical fidelity and implementation
cost. Importantly, the second-order Taylor term $(x-a)^2/2$ depends only on
the fractional residual, enabling its computation with a 12-bit multiplier.
The anchor-point exponential $e^{a}$ is pre-computed and absorbed into the
layer-specific scaling factor during calibration, reducing the inference-time
computation to a single 12-bit multiplication.

Since the multiplier constitutes the primary arithmetic bottleneck, this
fusion enables both division and square-root operations to be implemented
within a single computation cycle. The 12-bit intermediate precision is used
only for internal accumulation to prevent error propagation, while the final
output is re-quantized to 8-bit before being forwarded to subsequent layers.


\begin{figure}[t]
    \centering
    \begin{subfigure}{0.48\columnwidth}
        \centering
        \includegraphics[width=\linewidth]{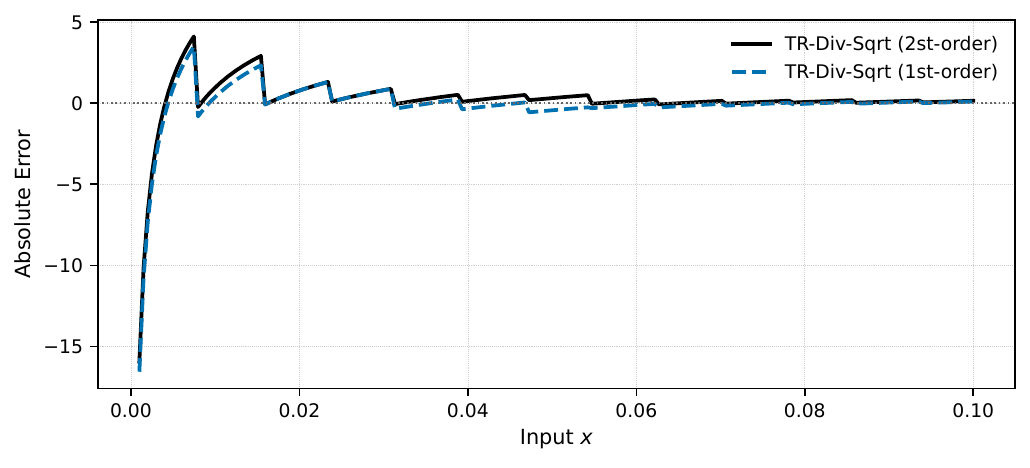}
        \caption{$[0.001,\,0.1]$, 8-bit}
    \end{subfigure}
    \hfill
    \begin{subfigure}{0.48\columnwidth}
        \centering
        \includegraphics[width=\linewidth]{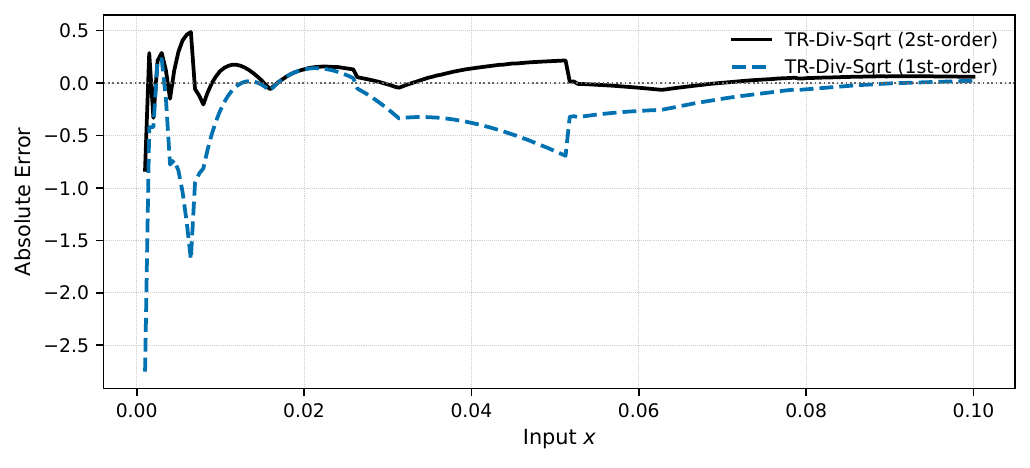}
        \caption{$[0.001,\,0.1]$, 12-bit}
    \end{subfigure}

    \vspace{0.5em}

    \begin{subfigure}{0.48\columnwidth}
        \centering
        \includegraphics[width=\linewidth]{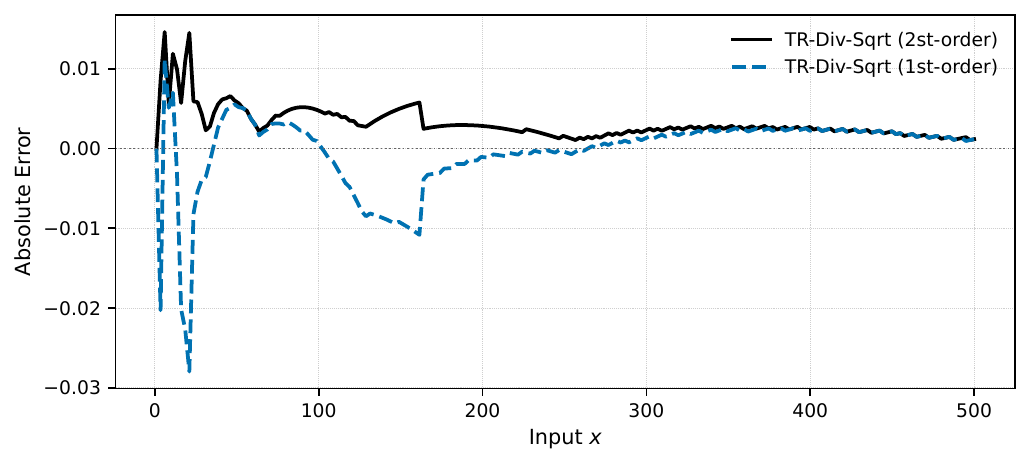}
        \caption{$[1,\,500]$}
    \end{subfigure}
    \hfill
    \begin{subfigure}{0.48\columnwidth}
        \centering
        \includegraphics[width=\linewidth]{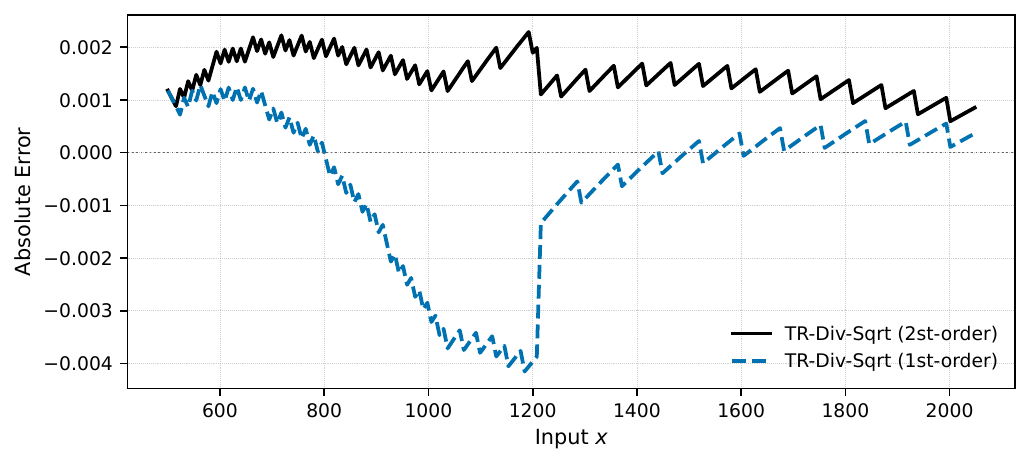}
        \caption{$[500,\,2048]$}
    \end{subfigure}

    \caption{
    Division noise comparison between the proposed method and a
    low-cost comparable solution (SOLE) across different input ranges
    and bit-precisions. Noise is measured as absolute error with respect
    to a high-precision reference.
    }
    \label{fig:division_noise_sole_multi}
\end{figure}

\section{Statistical Analysis of Affine Parameters}

\subsection{Gamma-like distribution of learned scales}
A key challenge in the quantization of normalization layers arises from the distribution of the learned affine parameters $\gamma$. Unlike weights in convolutional or linear layers, which often follow near-Gaussian statistics, the scale parameters in Transformer normalization layers are strictly
positive and frequently heavy-tailed. As a result, a small number of large values can dominate the dynamic range.

\begin{figure*}[t]
    \centering

    \begin{subfigure}{0.495\textwidth}
        \centering
        \includegraphics[width=\linewidth]{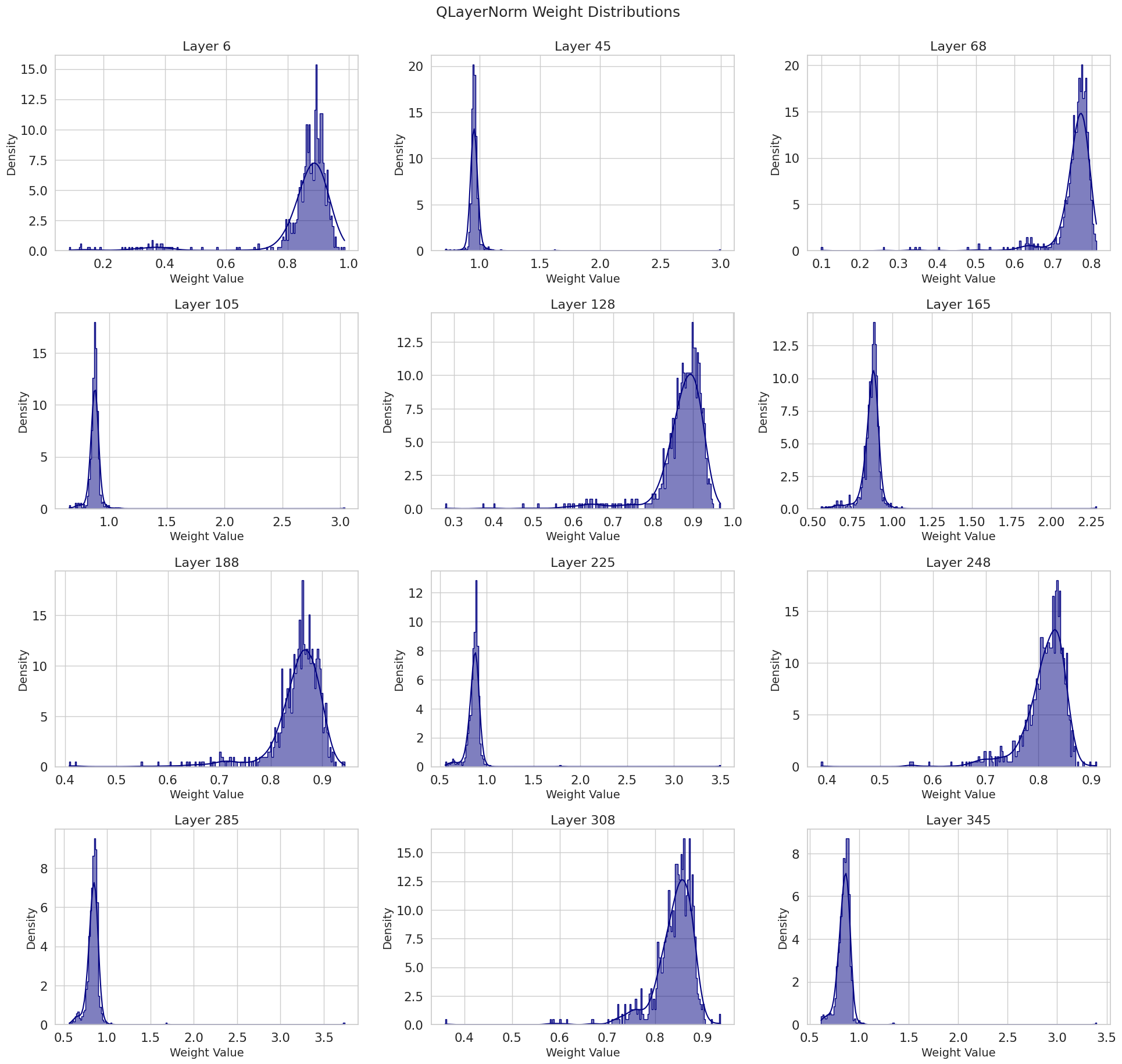}
        \caption{Original $\gamma$ matrices before min--max quantization (FP).}
    \end{subfigure}
    \begin{subfigure}{0.495\textwidth}
        \centering
        \includegraphics[width=\linewidth]{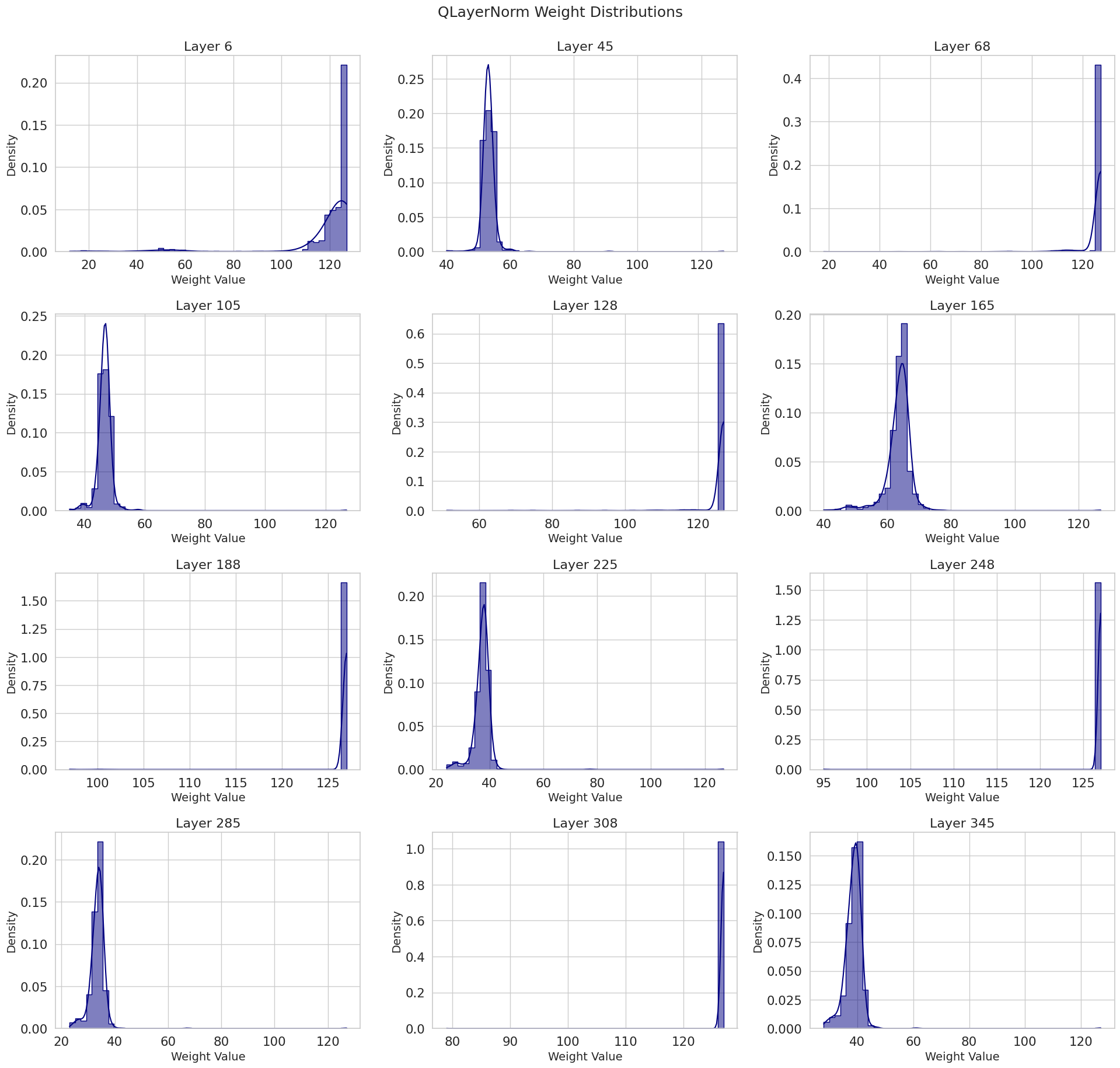}
        \caption{After min--max quantization (12-bit), showing degraded scaling.}
    \end{subfigure}

    \vspace{0.25em}

    \begin{subfigure}{0.495\textwidth}
        \centering
        \includegraphics[width=\linewidth]{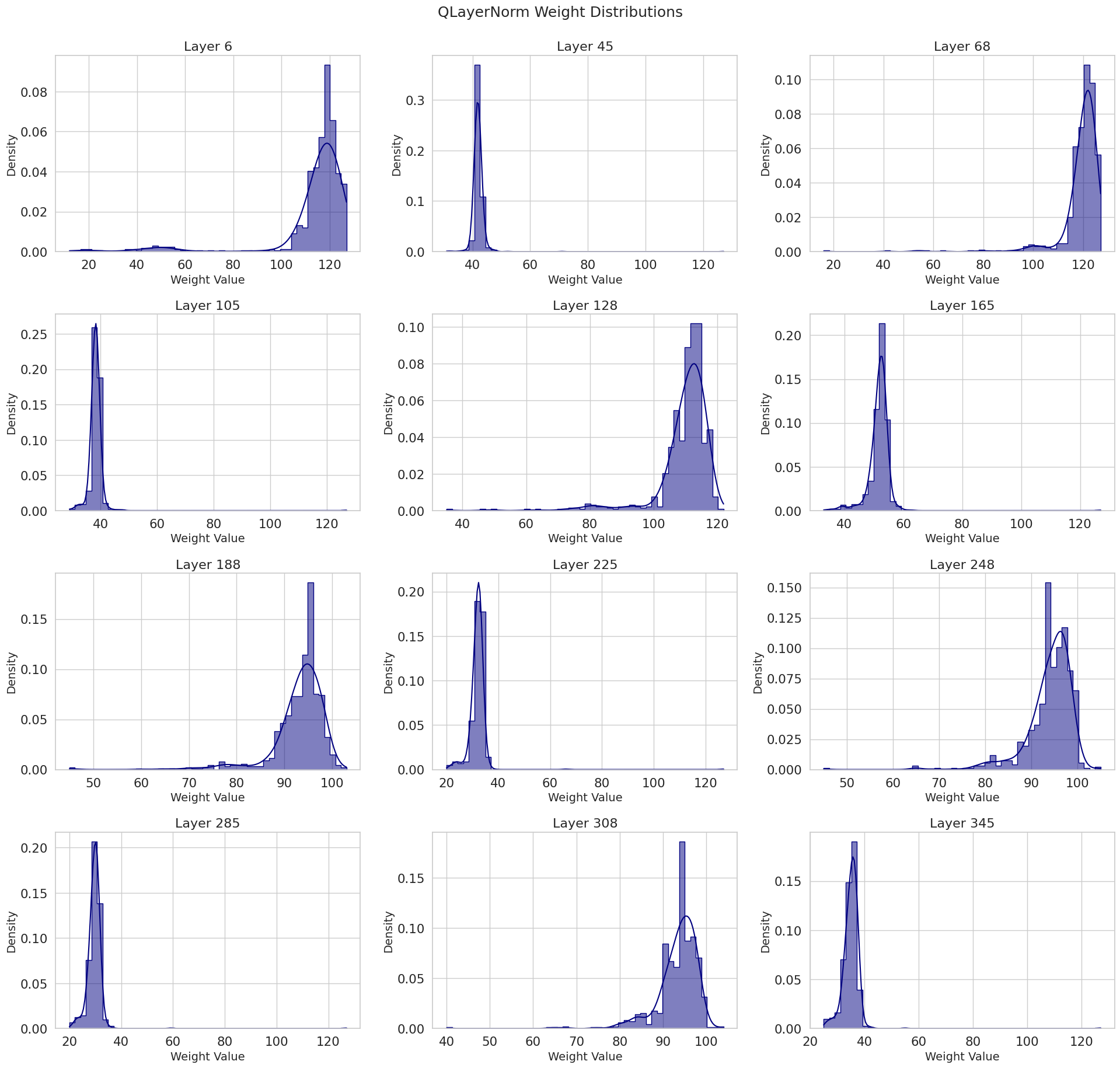}
        \caption{After relaxation, prior to min--max quantization (12-bit).}
    \end{subfigure}
    \begin{subfigure}{0.495\textwidth}
        \centering
        \includegraphics[width=\linewidth]{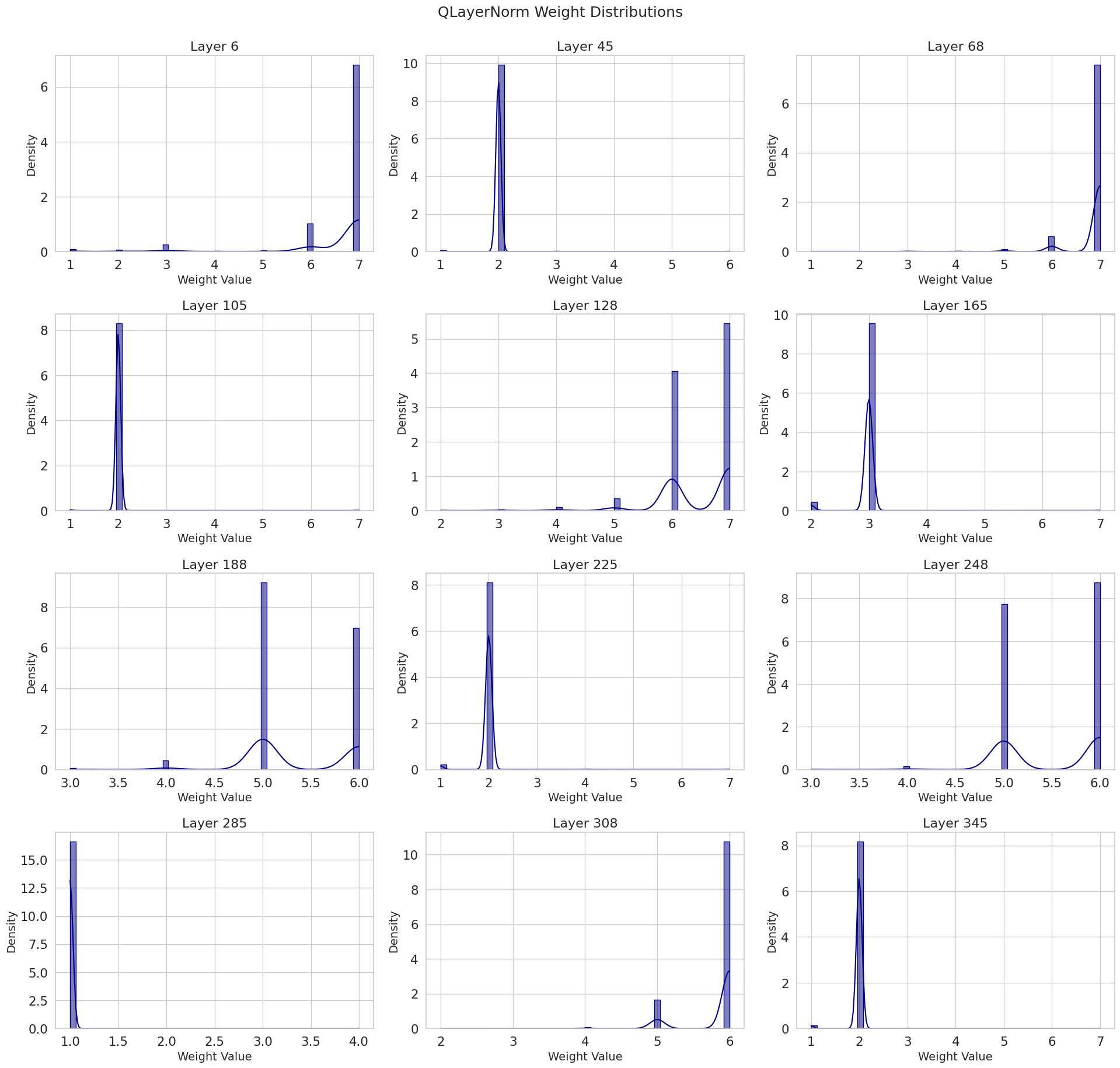}
        \caption{After relaxation, prior to min--max quantization (4-bit).}
    \end{subfigure}

    \caption{
    Effect of relaxation on \layernorm $\gamma$ parameters under min-max
    quantization. The top row shows the original $\gamma$ matrices before
    quantization and the corresponding 12-bit min-max quantized result,
    where scale distortion is observed. The bottom row illustrates the same matrices after relaxation, prior to quantization: the relaxed parameters remain closer to the original, and even under a 4-bit representation
    exhibit improved stability compared to the unrelaxed 12-bit case.
    }
    \label{fig:gamma_relaxation_effect}
\end{figure*}

Fig.~\ref{fig:gamma_relaxation_effect} illustrates the effect of these outliers under naive min-max quantization. In several layers, the dynamic range is dictated by a few extreme values, causing the majority of $\gamma$ parameters to collapse into the same quantization bin. This behavior significantly degrades the effective expressivity of the normalization layer.

\subsection{Distribution-preserving scale optimization}

\paragraph{Variance-normalized line search.}
Learned affine parameters ($\gamma$) often exhibit heavy-tailed distributions in which large-magnitude values are not noise, but play a functional role in model expressivity. Standard clipping or percentile-based schemes are therefore suboptimal, as they may artificially compress the distribution and force pre-activations into a rigid unit-variance regime.

To preserve the original distributional characteristics, we perform an offline line search for an optimal scaling factor $s \in [0.5, 2.0]$ during calibration. For each affine layer, we select $s$ by minimizing a variance-normalized RMSE objective:
\[
\mathcal{L}(s) =
\frac{\mathrm{RMSE}\!\left(\gamma,\,
\mathrm{dequant}(\mathrm{quant}(\gamma \cdot s))\right)}
{\mathrm{Var}(\gamma)} .
\]
Normalizing the reconstruction error by the original parameter variance penalizes scales that would collapse the distribution. This objective favors quantized representations that retain the fluctuation structure of the FP32 baseline, thereby preserving the relative importance of high-magnitude channels.

\paragraph{Empirical search range.}
Across all evaluated models, the optimal scale consistently lies within the $[0.5, 2.0]$ interval. Restricting the search to this range allows the TR-PTQ pipeline to adapt to the distinct statistical profiles of different architectures (e.g., higher variance in DeiT versus more localized scales in
Swin) without per-layer heuristics or training-aware fine-tuning.

\paragraph{Mitigating gamma-induced degradation.}
\begin{table}[t]
    \centering
    \small
    \renewcommand{\arraystretch}{1.2}
    \setlength{\tabcolsep}{4pt} 

    \caption{Vision model performance (Top-1 Acc. \%) under 4-bit weight and 12-bit activation quantization.}
    \label{tab:w4a12_vision_transposed}
    \begin{tabular}{@{}lcccccc@{}}
        \toprule
        \textbf{Model} & DeiT-T & DeiT-S & DeiT-B & Swin-T & Swin-S & Swin-B \\
        \midrule
        \textbf{Acc. (\%)} & 70.71 & 78.67 & 81.03 & 78.48 & 82.60 & 82.79 \\
        \bottomrule
    \end{tabular}

    \vspace{2em}

    \caption{GLUE benchmark results under 4-bit weight and 12-bit activation quantization.}
    \label{tab:w4a12_glue_transposed}
    \begin{tabular}{@{}lcccccc@{}}
        \toprule
        \textbf{Task} & CoLA & SST-2 & QQP & MNLI & QNLI & RTE \\
        \midrule
        \textbf{Metric} & 50.73 & 92.66 & 89.71 & 83.03 & 88.14 & 71.84 \\
        \bottomrule
    \end{tabular}
\end{table}

To assess the sensitivity of the affine parameters, we conduct an ablation in which $\gamma$ is quantized to 4-bit precision while all other parameters remain in floating point. Under naive min-max quantization, the heavy-tailed distribution of $\gamma$ leads to severe resolution loss for most channels,
resulting in a dramatic degradation in model accuracy.

Applying the proposed distribution-preserving line search restores functional behavior. As reported in Tables~\ref{tab:w4a12_glue_transposed} and \ref{tab:w4a12_vision_transposed}, the optimized models recover to within 1--2\% of their
FP32 baselines. For example, Swin-T recovers to 78.48\% Top-1 accuracy. This result indicates that scale optimization is not merely a refinement step, but a prerequisite for stable low-precision quantization of normalization layers in integer-only Transformer models.

\end{document}

%% file: newcommands/units.tex
\newcommand{\mm}{\,\si{\mm}\xspace}

\newcommand{\m}{\,\si{\m}\xspace}

\newcommand{\nm}{\,\si{\nm}\xspace}

\newcommand{\um}{\,\si{\um}\xspace}

\newcommand{\s}{\,\si{\s}\xspace}
\newcommand{\ps}{\,\si{\ps}\xspace}
\newcommand{\ns}{\,\si{\ns}\xspace}
\newcommand{\us}{\,\si{\us}\xspace}
\newcommand{\ms}{\,\si{\ms}\xspace}

\newcommand{\bit}{\,\si{\bit}\xspace}

\newcommand{\Hz}{\,\si{\Hz}\xspace}
\newcommand{\kHz}{\,\si{\kHz}\xspace}
\newcommand{\MHz}{\,\si{\MHz}\xspace}
\newcommand{\GHz}{\,\si{\GHz}\xspace}



%% file: newcommands/macros.tex
\renewcommand{\eqref}[1]{(\ref{#1})}
\newcommand{\secref}[1]{\mbox{Section~\ref{#1}}}

\newcommand{\figref}[1]{\mbox{Figure~\ref{#1}}}

\newcommand{\tblref}[1]{\mbox{Table~\ref{#1}}}

\definecolor{applegreen}{rgb}{0.55, 0.71, 0.0}

\usepackage{environ}
\newcounter{charcount}
\newcounter{charlim}
\NewEnviron{countenv}[2][]{%
  \setcounter{charcount}{0}%
  \expandafter\countem\BODY\relax\EOE
  \relax%
  \ifnum\value{charcount}<\value{charlim}\relax
      \begin{#2}\BODY\end{#2}
  \else
      \begin{#2}#1 \BODY  
         \color{red}\thecharcount{} letters in the #2 exceeds
         \thecharlim{} character limit. 
         Use the ``charlim'' parameter to change the character limit. \color{black}
       \end{#2} 
  \fi
}

\long\def\countem#1#2\EOE{%
  \stepcounter{charcount}%
  \ifx\relax#2
    \def\next{\relax}%
  \else
    \def\next{\countem#2\EOE}%
  \fi
  \expandafter\next%
}



%% file: newcommands/this-glossary.tex
\newacronym{cnn}{CNN}{convolutional neural network}

\newacronym{vit}{ViT}{vision transformer}
    \newcommand{\vit}{\gls{vit}\xspace}
    
\newacronym{QAT}{QAT}{quantization-aware training}
     
\newacronym{PTQ}{PTQ}{post-training quantization}
     \newcommand{\PTQ}{\gls{PTQ}\xspace}
\newcommand{\softmax}{SoftMax\xspace}
\newacronym{layernorm}{LayerNorm}{layer normalization}
     \newcommand{\layernorm}{\gls{layernorm}\xspace}
\newacronym{GELU}{GELU}{Gaussian Error Linear Unit}
     \newcommand{\GELU}{\gls{GELU}\xspace}
\newacronym{ppa}{PPA}{power, performance, and area}
      
\newacronym{lut}{LUT}{lookup table}
     \newcommand{\lut}{\gls{lut}\xspace}
          
\newacronym{mlp}{MLP}{multi-layer perceptron}
     
\newcommand{\tsoft}{TR-SoftMax\xspace}
\newcommand{\texp}{TR-$\exp$\xspace}
\newcommand{\tln}{TR-$\ln$\xspace}

\newcommand{\tgelu}{TR-GELU\xspace}
\newcommand{\tnorm}{TR-Norm\xspace}
\newcommand{\tvit}{TR-PTQ\xspace}
\newacronym{ai}{AI}{artificial-intelligence}

\newacronym{sota}{SoTA}{state-of-the-art}

\newacronym{msb}{MSB}{most-significant bit}

\newacronym{lsb}{LSB}{least-significant bit}

\newacronym{cdf}{CDF}{cumulative distribution function}
    \newcommand{\cdf}{\gls{cdf}\xspace}
\newacronym{fp}{FP32}{32-bit floating point}


%% file: ref/references.bib
@article{kim2021ibert,
  author       = {Sehoon Kim and
                  Amir Gholami and
                  Zhewei Yao and
                  Michael W. Mahoney and
                  Kurt Keutzer},
  title        = {{I-BERT:} Integer-only {BERT} Quantization},
  journal      = {CoRR},
  volume       = {abs/2101.01321},
  year         = {2021},
  url          = {https://arxiv.org/abs/2101.01321},
  eprinttype    = {arXiv},
  eprint       = {2101.01321},
  bibsource    = {dblp computer science bibliography, https://dblp.org}
}

@inproceedings{li2023vit,
  title={{I-ViT: Integer-only quantization for efficient vision transformer inference}},
  author={Li, Zhikai and Gu, Qingyi},
  booktitle={Proceedings of the IEEE/CVF International Conference on Computer Vision},
  pages={17065--17075},
  year={2023}
}

@article{zhang2023practical,
  title={{Practical edge kernels for integer-only vision transformers under post-training quantization}},
  author={Zhang, Zining and He, Bingsheng and Zhang, Zhenjie},
  journal={Proceedings of Machine Learning and Systems},
  volume={5},
  pages={35--47},
  year={2023}
}

@INPROCEEDINGS{softermax,
  author={Stevens, Jacob R. and Venkatesan, Rangharajan and Dai, Steve and Khailany, Brucek and Raghunathan, Anand},
  booktitle={2021 58th ACM/IEEE Design Automation Conference (DAC)}, 
  title={{Softermax: Hardware/Software Co-Design of an Efficient Softmax for Transformers}}, 
  year={2021},
  volume={},
  number={},
  pages={469-474},
  doi={10.1109/DAC18074.2021.9586134}}

@inproceedings{touvron2021training,
  title = {{Training Data-Efficient Image Transformers \& Distillation Through Attention}},
  author = {Touvron, Hugo and Cord, Matthieu and Douze, Matthijs and Massa, Francisco and Sablayrolles, Alexandre and J{\'e}gou, Herv{\'e}},
  booktitle = {Proceedings of the IEEE Conference on Computer Vision and Pattern Recognition (CVPR)},
  pages = {10347--10357},
  year = {2021}
}

@inproceedings{liu2021swin,
  title={{Swin transformer: Hierarchical vision transformer using shifted windows}},
  author={Liu, Ze and Lin, Yutong and Cao, Yue and Hu, Han and Wei, Yixuan and Zhang, Zheng and Lin, Stephen and Guo, Baining},
  booktitle={Proceedings of the IEEE/CVF international conference on computer vision},
  pages={10012--10022},
  year={2021}
}

@article{lin2021fqvit,
  title={{FQ-ViT: Post-training quantization for fully quantized vision transformer}},
  author={Lin, Yang and Zhang, Tianyu and Sun, Peiqin and Li, Zheng and Zhou, Shuchang},
  journal={arXiv preprint arXiv:2111.13824},
  year={2021}
}

@misc{crandall2001computational,
  title={{Prime Numbers: A computational perspective}},
  author={Crandall, Richard and Pomerance, Carl},
  year={2001},
  publisher={Springer, New York}
}

@article{gelu2016,
  author       = {Dan Hendrycks and
                  Kevin Gimpel},
  title        = {Bridging Nonlinearities and Stochastic Regularizers with Gaussian
                  Error Linear Units},
  journal      = {CoRR},
  volume       = {abs/1606.08415},
  year         = {2016},
  url          = {http://arxiv.org/abs/1606.08415},
  eprinttype    = {arXiv},
  eprint       = {1606.08415},
  bibsource    = {dblp computer science bibliography, https://dblp.org}
}

@ARTICLE{fastertransformer,
  author={Zhang, Jiahuan and Liu, Hengzhen and Zhang, Yi and Li, Menghan and Zhan, Zongqian},
  journal={IEEE Journal of Selected Topics in Applied Earth Observations and Remote Sensing}, 
  title={{Faster Transformer-DS: Multiscale Vehicle Detection of Remote-Sensing Images Based on Transformer and Distance-Scale Loss}}, 
  year={2024},
  volume={17},
  number={},
  pages={1961-1975},
  doi={10.1109/JSTARS.2023.3335283}}

@inproceedings{deng2009imagenet,
  title={{ImageNet: A large-scale hierarchical image database}},
  author={Deng, Jia and Dong, Wei and Socher, Richard and Li, Li-Jia and Li, Kai and Fei-Fei, Li},
  booktitle={2009 IEEE conference on computer vision and pattern recognition},
  pages={248--255},
  year={2009},
  organization={IEEE}
}

@article{ba2016layer,
  title={{Layer normalization}},
  author={Ba, Jimmy Lei and Kiros, Jamie Ryan and Hinton, Geoffrey E},
  journal={arXiv preprint arXiv:1607.06450},
  year={2016}
}

@INPROCEEDINGS{SOLE2023_ICCAD,
  author={Wang, Wenxun and Zhou, Shuchang and Sun, Wenyu and Sun, Peiqin and Liu, Yongpan},
  booktitle={2023 IEEE/ACM International Conference on Computer Aided Design (ICCAD)}, 
  title={{SOLE: Hardware-Software Co-design of Softmax and LayerNorm for Efficient Transformer Inference}}, 
  year={2023},
  volume={},
  number={},
  pages={1-9},
  doi={10.1109/ICCAD57390.2023.10323725}}

@INPROCEEDINGS{Park2025_CSCC, 
  author={Park, Junseok and Rhee, Chae Eun},
  booktitle={2025 International Technical Conference on Circuits/Systems, Computers, and Communications (ITC-CSCC)}, 
  title={{High-Precision Softmax Division without Multipliers or Look-Up Tables}}, 
  year={2025},
  volume={},
  number={},
  pages={1-3},
  doi={10.1109/ITC-CSCC66376.2025.11137704}}

@INPROCEEDINGS{ARNorm2025_CSCC,
  author={Kim, Sunyeop and Rhee, Chae Eun},
  booktitle={2025 International Technical Conference on Circuits/Systems, Computers, and Communications (ITC-CSCC)}, 
  title={{ARNorm: Hardware-Efficient Normalization for Lightweight Edge Models}}, 
  year={2025},
  volume={},
  number={},
  pages={1-3},
  doi={10.1109/ITC-CSCC66376.2025.11137703}}

@INPROCEEDINGS{QUARK2025_ICCAD,
  author={Zhao, Zhixiong and Li, Haomin and Liu, Fangxin and Lu, Yuncheng and Wang, Zongwu and Yang, Tao and Jiang, Li and Guan, Haibing},
  booktitle={2025 IEEE/ACM International Conference On Computer Aided Design (ICCAD)}, 
  title={{QUARK: Quantization-Enabled Circuit Sharing for Transformer Acceleration by Exploiting Common Patterns in Nonlinear Operations}}, 
  year={2025},
  volume={},
  number={},
  pages={1-9},
  doi={10.1109/ICCAD66269.2025.11240736}}

@article{frantar2022gptq,
  title={Gptq: Accurate post-training quantization for generative pre-trained transformers},
  author={Frantar, Elias and Ashkboos, Saleh and Hoefler, Torsten and Alistarh, Dan},
  journal={arXiv preprint arXiv:2210.17323},
  year={2022}
}

@InProceedings{pmlr-v202-xiao23c,
  title = 	 {{S}mooth{Q}uant: Accurate and Efficient Post-Training Quantization for Large Language Models},
  author =       {Xiao, Guangxuan and Lin, Ji and Seznec, Mickael and Wu, Hao and Demouth, Julien and Han, Song},
  booktitle = 	 {Proceedings of the 40th International Conference on Machine Learning},
  pages = 	 {38087--38099},
  year = 	 {2023},
  editor = 	 {Krause, Andreas and Brunskill, Emma and Cho, Kyunghyun and Engelhardt, Barbara and Sabato, Sivan and Scarlett, Jonathan},
  volume = 	 {202},
  series = 	 {Proceedings of Machine Learning Research},
  month = 	 {23--29 Jul},
  publisher =    {PMLR},
  url = 	 {https://proceedings.mlr.press/v202/xiao23c.html}
}

@INPROCEEDINGS{11270824,
  author={Sun, Tianyi and Ma, Tuo and Liu, Jiali and Li, Zhiwei and Li, Qingjiang and Wang, Yinan and Liu, Haijun and Liu, Sen},
  booktitle={2025 32nd IEEE International Conference on Electronics, Circuits and Systems (ICECS)}, 
  title={{Integer Quantization of Nonlinear Operations towards Hardware-Friendly ViTs}}, 
  year={2025},
  volume={},
  number={},
  pages={1-4},
  doi={10.1109/ICECS66544.2025.11270824}}

@misc{wu2020easyquantposttrainingquantizationscale,
      title={EasyQuant: Post-training Quantization via Scale Optimization}, 
      author={Di Wu and Qi Tang and Yongle Zhao and Ming Zhang and Ying Fu and Debing Zhang},
      year={2020},
      eprint={2006.16669},
      archivePrefix={arXiv},
      primaryClass={cs.CV},
      url={https://arxiv.org/abs/2006.16669}, 
}
